\documentclass[10pt,twocolumn,letterpaper]{article}

\usepackage{3dv}
\usepackage{times}
\usepackage{epsfig}
\usepackage{graphicx}
\usepackage{amsmath}
\usepackage{amssymb}

\usepackage{adjustbox} 
\usepackage{multirow}
\usepackage{booktabs}
\usepackage{array}
\usepackage{caption}
\usepackage{chngcntr}

\DeclareMathOperator*{\argmin}{argmin}
\newcolumntype{P}[1]{>{\centering\arraybackslash}p{#1}}

\usepackage[pagebackref=true,breaklinks=true,colorlinks,bookmarks=false]{hyperref}

\threedvfinalcopy 

\begin{document}

\title{Inertial Guided Uncertainty Estimation of Feature Correspondence\\in Visual-Inertial Odometry/SLAM}

\author{Seongwook Yoon$^*$\\
Korea University\\
Seoul, Republic of Korea\\
{\tt\small swyoon@mpeg.korea.ac.kr}
\and
Jaehyun Kim$^*$\\
Korea University\\
Seoul, Republic of Korea\\
{\tt\small jhkim@mpeg.korea.ac.kr}
\and
Sanghoon Sull\\
Korea University\\
Seoul, Republic of Korea\\
{\tt\small sull@korea.ac.kr}
}

\maketitle
\def\thefootnote{*}\footnotetext{These authors contributed equally to this work}

\begin{abstract}

Visual odometry and Simultaneous Localization And Mapping (SLAM) has been studied as one of the most important tasks in the areas of computer vision and robotics, to contribute to autonomous navigation and augmented reality systems. In case of feature-based odometry/SLAM, a moving visual sensor observes a set of 3D points from different viewpoints, correspondences between the projected 2D points in each image are usually established by feature tracking and matching. However, since the corresponding point could be erroneous and noisy, reliable uncertainty estimation can improve the accuracy of odometry/SLAM methods. In addition, inertial measurement unit is utilized to aid the visual sensor in terms of Visual-Inertial fusion. In this paper, we propose a method to estimate the uncertainty of feature correspondence using an inertial guidance robust to image degradation caused by motion blur, illumination change and occlusion. Modeling a guidance distribution to sample possible correspondence, we fit the distribution to an energy function based on image error, yielding more robust uncertainty than conventional methods. We also demonstrate the feasibility of our approach by incorporating it into one of recent visual-inertial odometry/SLAM algorithms for public datasets.

\end{abstract}

\section{Introduction}

Visual odometry has been an important research topic in both computer vision and robotics community. Visual odometry is to estimate time-series of pose (position and orientation) of moving visual sensors, mostly RGB cameras due to efficiency and cost. There are various applications such as self-navigating robot, autonomous driving and virtual/augmented reality. Basically, it utilizes consecutive images to find correspondence between the images and the stationary world. Visual odometry can be extended to simultaneous localization and mapping (SLAM) ~\cite{engel2014lsd,klein2007parallel,mur2017orb} which builds a 3D world map. It can also utilize other sensors ~\cite{Schops_2019_CVPR,liang2016visual,turan2017magnetic} such as inertial measurement unit (IMU) ~\cite{mourikis2007multi, campos2021orb, qin2018vins} and LIDAR ~\cite{jiang2019simultaneous, xu2018slam, shin2018direct}. Nonetheless, finding the correspondences in images taken from visual sensors is still one of the most important processes in visual odometry algorithms and applications.

One of the most efficient ways to find the correspondence is considering feature points which are interesting or useful points in both image and world. Two approaches are utilized to find the correspondence based on the feature points: The feature tracking approach ~\cite{lucas1981iterative,bouguet2001pyramidal,hwangbo2009inertial} finds the image point most similar to a reference feature point and the feature matching approach ~\cite{lowe2004distinctive,bay2006surf,rublee2011orb} matches a pair of two different feature points in two respective images. On the other hand, there are also various ways to find dense correspondences between two images ~\cite{horn1981determining,brox2010large,dosovitskiy2015flownet}.

Most of the above methods find the correspondence based on visual features such as pixel values and image structure to measure local similarities. Then, it is assumed that such visual correspondence ensures true correspondence which allows accurate localization and mapping. The true correspondence would be defined by the exact projection from the same stationary point in 3D world. However, it is difficult to find the same point using images, when the visual features are degraded by illumination change, occlusion and motion blur. Even if the visual correspondence is well established as the local maximum of visual similarity, it can differ from the true correspondence. 

In trivial cases, fortunately, wrong correspondences can be eliminated by outlier detection ~\cite{fischler1981random, torr2000mlesac, chum2005matching, hodge2004survey} or failure check algorithms~\cite{mur2017orb}. However, unfortunately, it is difficult to remove the correspondences with relatively small errors increasing their uncertainties. It is believed that employing a large number of correspondences or even dense correspondences can increase the accuracy of the results over time. Otherwise, the uncertainty should be reliably estimated to assure more accurate odometry.

There exist various works on the uncertainty estimation of correspondence between feature points. The earlier methods~\cite{singh1989estimation, nickels2002estimating} are based on image errors between a patch centered at a reference point and other patches centered at the points on a square grid. They calculate uncertainty from an energy function derived from a scaled patch error surface using two slightly different heuristic conditions. Rather than the uncertainty defined above, the later approaches concentrate on Kanade–Lucas–Tomasi (KLT) feature tracking algorithms~\cite{lucas1981iterative}. 

\cite{sheorey2014uncertainty} also observes the patch errors to analyze how a noisy reference point results in different KLT tracking results. Recently, \cite{iuan2017uncertainty} calculates an analogous uncertainty without such observations by linear analysis of the sensitivity of the KLT tracking algorithm. In this paper, we define the uncertainty of visual feature correspondence as the probabilistic difference between visual correspondence and true correspondence, regardless of specific tracking or matching algorithm.

As if RGB camera is mainly utilized as one of the most efficient and affordable sensor for odometry/SLAM, an IMU is also adopted due to its cost effectiveness and suitability. It can be localized simply by integrating its acceleration and angular velocity over time, but it is vulnerable to the accumulated error caused by initial velocity and biases. Visual-inertial odometry (VIO) fuses RGB camera and IMU to compensate for their weaknesses. 

For example, the velocity and biases for IMU integration are jointly determined with other state estimation such as bundle adjustment. At the same time, the inertial odometry can give useful guidance to finding correspondence and rejecting outliers, because the inertial odometry is nearly independent with visual degradation of incoming images. Likewise, in this paper, we suggest that the uncertainty estimation of visual correspondence also takes advantage of the inertial odometry. 

Therefore, we propose a novel uncertainty estimation method for feature-based VIO/SLAM. Since we define the uncertainty as a distribution of true correspondence, the uncertainty not only yields a covariance matrix but also changes the actual corresponding point with the mean of the distribution. Also, the point is that the true correspondence should be predicted only by preceding clues, such as the latest results of inertial pre-integration and earlier image process. To this end, we sample multiple guidance points using the inertial pose estimate and the probable values of depths and generate their respective guidance distributions each of which is a prior distribution of possible corresponding points. Then, we sample multiple sets of possible corresponding points to minimize Kullback–Leibler divergence between the prior distribution and a distribution derived by image errors on the possible points. Then, we extend these point-level uncertainty to a number of sequential correspondences in a sequence of images.


\section{Related work}

\textbf{Odometry/SLAM:} Most of the odometry and SLAM algorithms are divided into two parts, the front-end for sensor measurement process and the back-end for state estimation. Conventionally, the front-end for a visual sensor uses feature-based algorithms (e.g. feature tracking and matching) and the back-end utilizes either Kalman Filter~\cite{davison2007monoslam,mourikis2007multi} or bundle adjustment~\cite{klein2007parallel,mur2017orb,qin2018vins}. Though either or both parts can be replaced partially with deep neural networks~\cite{detone2018superpoint, sarlin2020superglue,yang2020d3vo}, geometry based approaches still show competitive performance and better efficiency.

\textbf{Visual-inertial fusion:} Combining a visual sensor with an inertial sensor is beneficial in odometry/SLAM. For example, since it is difficult to estimate the absolute scale of depth and motion using a monocular camera, an IMU is used to solve the scale problem and enable robust rotation estimation using the gravity~\cite{qin2018vins}.

\textbf{Feature tracking/matching:} A feature point is defined as the most distinguishable point in a local image area, such as a corner and blob. There are two major approaches to finding the corresponding point of a given feature point in another image. Feature tracking tries to find the corresponding point among all the points in a search range. On the other hand, feature matching identifies the corresponding point among the feature points detected in a search range. While either method can be chosen considering the pros and cons, both approaches are good at finding correspondence in consecutive images. Also, in VIO/SLAM, an IMU can provide better search range or outlier rejection~\cite{hwangbo2009inertial,troiani20142,masiero2016improved}.

\textbf{Inertial odometry:} Inertial Odometry integrates acceleration and angular velocity measurements from IMU. Since the integration causes error accumulation, it is important to estimate bias and initial velocity ~\cite{harle2013survey}. Thus, a number of researches try to estimate those values using Extended Kalman Filter~\cite{foxlin2005pedestrian, jimenez2010indoor}, or deep learning methods~\cite{yan2019ronin, chen2018ionet}. Fortunately, in VIO, the problem of error accumulation is relieved, because the latest bias and velocity can be estimated consecutively from the back-end. Thus, the simple integration can give quite accurate poses in a short period before optimization, or even an image analysis in the front-end for the purpose of either pre-integration in the front-end or forward propagation.

\section{Overview}

\begin{figure*}[t]\centering
  \includegraphics[height=4.5cm]{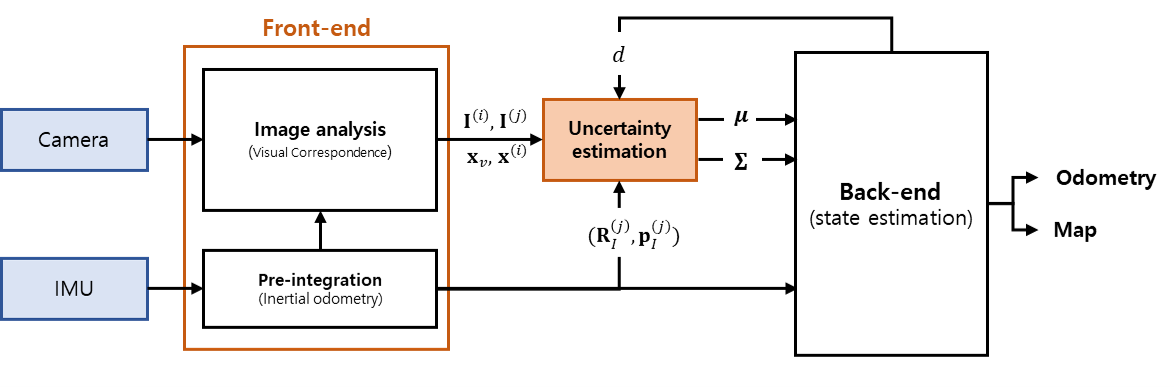}
  \caption{Overview of the proposed inertial guided uncertainty estimation in Visual-Inertial odometry/SLAM.}
\label{fig:overview}
\end{figure*}

The definition of correspondence for uncertainty estimation is basically equivalent to the following general definition. Provided that a 3D point is projected onto a reference point $\boldsymbol{\mathrm{x}}^{(i)}$ in a reference image $\boldsymbol{\mathrm{I}}^{(i)}$, the true corresponding point $\boldsymbol{\mathrm{x}}^{(j)}$ in target image $\boldsymbol{\mathrm{I}}^{(j)}$ is a projection of the identical 3D point. Obviously, the corresponding point can be determined by its depth $d$ at the reference time $i$ and true pose $(\boldsymbol{\mathrm{R}}^{(j)}, \boldsymbol{\mathrm{p}}^{(j)})$ at time $j$.

As shown in Fig.~\ref{fig:overview}, to clarify our objective, we aim to predict more probable corresponding point with uncertainty using only accessible information before the overall state estimation in the back-end. Thus, preceding estimates such as visual correspondence and inertial odometry through the pre-integration could be a good source of prior information.

To this end, the true pose assumed to be substituted with the inertial pose estimate $(\boldsymbol{\mathrm{R}}_{I}^{(j)}$ and $\boldsymbol{\mathrm{p}}_{I}^{(j)})$. Also, we can observe a visual corresponding point $\boldsymbol{\mathrm{x}}_{v}$ including the two images $\boldsymbol{\mathrm{I}}^{(i)}$ and $\boldsymbol{\mathrm{I}}^{(j)}$. Even if the depth $d$ is not estimated yet in the back-end, it can be approximated by triangulation using the visual correspondence and the inertial pose. Thus, we model a conditional distribution of correspondence $p(\boldsymbol{\mathrm{x}}^{(j)}| \boldsymbol{\mathrm{x}}^{(i)}, \boldsymbol{\mathrm{x}}_{v}, \boldsymbol{\mathrm{I}}^{(i)}, \boldsymbol{\mathrm{I}}^{(j)}, \boldsymbol{\mathrm{R}}_{I}^{(j)}, \boldsymbol{\mathrm{p}}_I^{(j)}, d)$. Then, we estimate both mean and covariance matrix of the distribution, whereas most existing uncertainty estimation methods compute covariance matrix only.

\begin{figure}[t]
\begin{center}
\includegraphics[width=\linewidth]{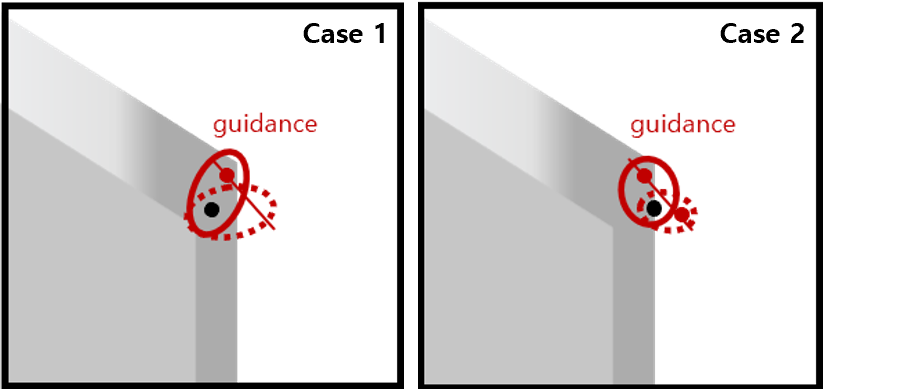}
\end{center}
   \caption{Different cases of guidance points (red dots) sampled from the epipolar line (red line segment) with different drift of visual corresponding points (block dots): Both drifted corner points near the blurred edge can be guided by the point projected robustly with the image degradation. The solid red ellipses represent the region of interest where possible corresponding points can be located. The dotted red ellipses represent failure cases by inaccurate guidance samples.}
\label{fig:figure_GuidancePoint}
\end{figure}

As mentioned above, a possible corresponding point can be calculated by a re-projection using the inertial pose estimate $(\boldsymbol{\mathrm{R}}_{I}^{(j)}, \boldsymbol{\mathrm{p}}_{I}^{(j)})$ and the depth $d$. We sample multiple guidance points, each of which is a hypothesis that the visual corresponding point may drift from each guidance point. Thus, a set of possible corresponding points can be sampled by the single sample of guidance point and the visual corresponding point. Then, we predict the most probable corresponding point upon the hypothesis using two kinds of energy function at the possible corresponding points. The hypothesis is described in Fig.~\ref{fig:figure_GuidancePoint}. There are two cases of drifted visual corresponding point. In both cases, we can assume that true corresponding point is in the vicinity, red ellipses in the figure, of the visual corresponding point and each sample of guidance point. The multiple samples of guidance point lead to multiple hypotheses.

However, the multiple hypotheses should be combined into one, because most of back-end algorithms assume a uni-modal distribution. In order to avoid too diverse hypotheses, we trust the visual correspondence more than each guidance point so that the true corresponding point can be predicted around it.

\section{Methods}
\label{section:Methods}
We present an uncertainty estimation algorithm for a single visual corresponding point in a target image for a given point in a reference image. The point-level procedure consists of 1) sampling guidance point from inertial odometry, 2) generating a guidance distribution using a distance-based energy function, 3) sampling possible corresponding points, 4) scaling another error-based energy function using the samples, and 5) point-level marginalization. In order to extend the point-level procedure to a window of consecutive images, we also propose methods for uncertainty propagation and image-level normalization.

\subsection{Inertial guidance sampling}
\label{section:Inertial guidance sampling}
Since inertial odometry gives the pose estimate of an incoming target image, a point $\boldsymbol{\mathrm{x}}^{(i)}$ in the reference image can be re-projected onto the target image with given depth $d$ as below:
\begin{equation}
\label{eqn:5}
\boldsymbol{\mathrm{x}}_{d}^{(j)}=\boldsymbol{\mathrm{\pi}}[\boldsymbol{\mathrm{R}}_{I}^{(i,j)}\boldsymbol{\mathrm{\pi}}^{-1}[\boldsymbol{\mathrm{x}}^{(i)},d]+\boldsymbol{\mathrm{p}}_{I}^{(i, j)}],
\end{equation}
where $\boldsymbol{\mathrm{\pi}}[\cdot]$ is the projection function from a 3D point in camera coordinates to an image point, and $\boldsymbol{\mathrm{\pi}}^{-1}[\cdot,d]$ is the back-projection function from an image point to a 3D point in camera coordinates with depth $d$. Also, the transformation between the two camera coordinates $(\boldsymbol{\mathrm{R}}_{I}^{(i,j)}, \boldsymbol{\mathrm{p}}_{I}^{(i,j)})$ is given by the difference between a target pose $(\boldsymbol{\mathrm{R}}_{I}^{(j)}, \boldsymbol{\mathrm{p}}_{I}^{(j)})$ from inertial odometry and its reference pose $(\boldsymbol{\mathrm{R}}^{(i)}, \boldsymbol{\mathrm{p}}^{(i)})$. Note that the reference pose also could be also obtained from inertial odometry before VIO computes the reference pose in the back-end.

Then, the guidance point $\boldsymbol{\mathrm{x}}_{g}$ is sampled by the re-projection as below:
\begin{equation}
\label{eqn:distirbution of guidance point}
p(\boldsymbol{\mathrm{x}}_{g}|\boldsymbol{\mathrm{R}}_{I}^{(i,j)}, \boldsymbol{\mathrm{p}}_{I}^{(i,j)})=\int{p(\boldsymbol{\mathrm{x}}_{g}|\boldsymbol{\mathrm{x}}_{d}^{(j)})p(d)\mathrm{d}d},
\end{equation}
where $p(d)$ is a sampling distribution of depth and $p(\boldsymbol{\mathrm{x}}_{g}|\boldsymbol{\mathrm{x}}_{d}^{(j)})$ is a clipping module to prevent wrong guidance. A clipped guidance point is relocated in maximum distance, defined as a hyper-parameter, from the visual corresponding poset to 0.76int $\boldsymbol{\mathrm{x}}_{v}$ so as to provide directional guidance. When the two points are close enough not to be clipped, it is reasonable to sample a guidance point on the epipolar line determined by the inertial pose estimate pose estimate.

Also, the sampled guidance point is still conditional to the inertial pose estimate between two images which implies all the correspondences between the two images are determined by the single inertial pose estimate. Thus, the noise of the inertial pose estimate should be considered by the frame-wise uncertainty normalization of the entire correspondences in the two images.

\subsection{Guidance distribution model}
\label{section:Guidance distribution model}

Since we hypothesize that the visual corresponding point drifts from the guidance point, the possible corresponding points are distributed around the line segment connecting the two points. Therefore, a distance-based energy function is defined by the weighted mean of the distances from the guidance point $\boldsymbol{\mathrm{x}}_{g}$ and the visual corresponding point $\boldsymbol{\mathrm{x}}_{v}$ as below:
\begin{equation}
\label{eqn:7}
\psi_{\text{d}}(\boldsymbol{\mathrm{x}};\boldsymbol{\mathrm{x}}_{v},\boldsymbol{\mathrm{x}}_{g})=\beta\left\Vert{\boldsymbol{\mathrm{x}}-\boldsymbol{\mathrm{x}}_{v}}\right\Vert_{1}+(1-\beta)\left\Vert{\boldsymbol{\mathrm{x}}-\boldsymbol{\mathrm{x}}_{g}}\right\Vert_{1},
\end{equation}
where $\beta\in[0, 1]$ is a weight parameter for balancing between the two points. If $\beta$ is close to 1, the prior become more dependent on the visual corresponding point. Note that the distribution is not symmetric due to the form of $l_1$ distances. We define the distances in the rotated axis aligned along the visual corresponding point and the guidance point. Also, a guidance distribution is defined by the distance-based energy function as below:
\begin{equation}
\label{eqn:8}
q(\boldsymbol{\mathrm{x}}|\boldsymbol{\mathrm{x}}_{v},\boldsymbol{\mathrm{x}}_{g})=\frac{\mathrm{exp}[-\alpha\psi_{\text{d}}(\boldsymbol{\mathrm{x}};\boldsymbol{\mathrm{x}}_{v},\boldsymbol{\mathrm{x}}_{g})]}{\mathrm{Z}_{\alpha,\beta}},
\end{equation}
where $\alpha$ is a scale parameter of the energy function and $\mathrm{Z}_{\alpha,\beta}$ is a normalization factor simply calculated by $\alpha$ and $\beta$. A sloped plateau is formed along the two points, which is the region of interest for possible corresponding points. 

The two parameters, $\alpha$ and $\beta$, are computed by the following two conditions about the slope of the plateau using the probability density values at the visual corresponding point $\boldsymbol{\mathrm{x}}_{v}$ and the guidance point $\boldsymbol{\mathrm{x}}_{g}$ as below:
\begin{equation}
\label{eqn:param_r}
q(\boldsymbol{\mathrm{x}}_{v}|\boldsymbol{\mathrm{x}}_{v},\boldsymbol{\mathrm{x}}_{g})=r*q(\boldsymbol{\mathrm{x}}_{g}|\boldsymbol{\mathrm{x}}_{v},\boldsymbol{\mathrm{x}}_{g}),
\end{equation}
where $r$ is a ratio parameter motivated by the slope of the distribution. As well as the plateau become flat and narrow as the two points get closer, it should decrease from a given maximum slope to 1. Also, we set the minimum of $r$ to 1 when the two points are identical, to guarantee the continuity of the distribution.

\subsection{Possible correspondence sampling}
\label{section:Possible correspondence sampling}
In addition to the guidance point and the visual corresponding point, we use the two images themselves. We assume that the negative log-likelihood, or energy, of the true corresponding point can be linearly approximated with an image statistics related to the visual correspondence. While the conventional approaches~\cite{singh1989estimation, nickels2002estimating} also assume the linear approximation, we argue the necessity to sample more likely points than a large square grid of the conventional approaches. Thus, we propose to sample points with high likelihoods with respect to the guidance distribution.

We want to sample points of which prior probabilities are larger than a certain threshold $c$. It is equivalent to sample points inside a contour of $c$-level set $\{\boldsymbol{\mathrm{x}}\in\mathbb{R}^2|p(\boldsymbol{\mathrm{x}}|\boldsymbol{\mathrm{x}}_{v},\boldsymbol{\mathrm{x}}_{g})=c\}$ of the guidance distribution. For simplicity, we find the smallest grid that contains the contour.

\begin{figure}[t]
\begin{center}
\includegraphics[scale=0.4]{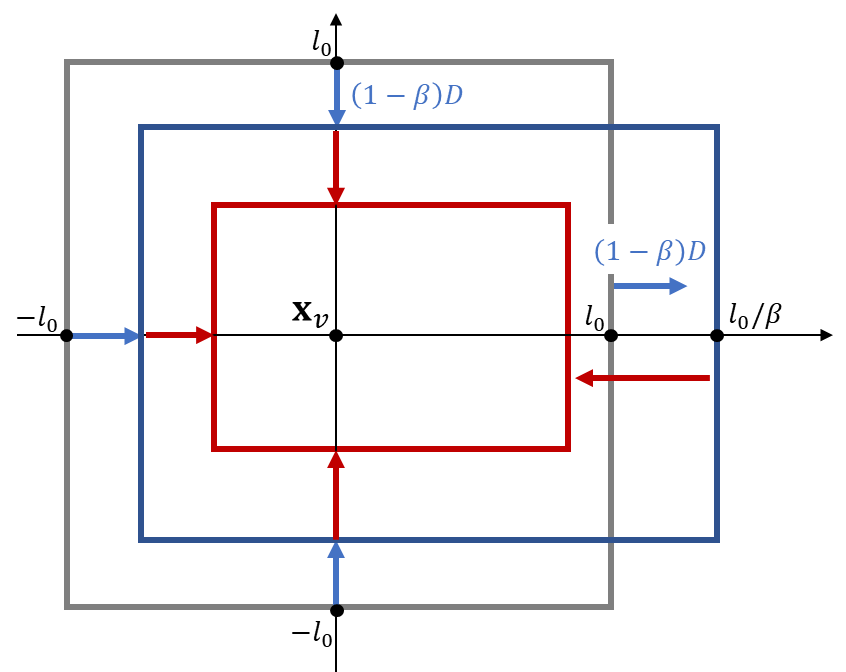}
\end{center}
   \caption{The behavior of the grid for sampling possible corresponding points: If the visual point and guidance point is identical, the square grid (gray box) is chosen. As long as the guidance point is inside of the square grid, the grid (blue box) moves and lengthens optimistically to the guidance point. Otherwise, the grid (red box) shrinks and shortens pessimistically to the guidance point.}
\label{fig:grid_sample}
\end{figure}

Fig.~\ref{fig:grid_sample} shows the behavior of the four outer sides of the grid with respect to the distance between the visual corresponding point and the guidance point. If these two points are identical, it becomes a square of which size is $l_{0}=-\frac{1}{\alpha}\log{c}$. Thus, we can determine the parameter $c$ backwards from a proper window size $l_{0}$. The three sides around the visual corresponding point get closer as $(1-\beta)D$. The remaining side around the guidance point get farther as $(1-\beta)D$, if the aspect ratio is smaller than $\beta/(2\beta-1)$. Otherwise, the aspect ratio is restrained and the grid shrinks. Such behavior is quite appropriate in practical sense. As long as the distance $D$ is clipped as described in Section~\ref{section:Inertial guidance sampling}, the grid balances the aspect ratio and the inner area not to be too biased by the guidance point. The grid tries to either contain a close guidance point, or abandon a relatively far guidance point. Then, a set of possible corresponding points $\boldsymbol{\mathrm{X}}$ are sampled from $N\times N$ grid, as below:
\begin{equation}
\boldsymbol{\mathrm{X}}=\{\boldsymbol{\mathrm{x}}_v+{\langle}wW/N-l,hH/N-l\rangle | 0{\leq}h,w{\leq}N\}
\label{eqn:gridset}
\end{equation}
where $h,w\in\mathbb{Z}$, $l=l_0-(1-\beta)D$, $H=2l$, and $W=2l_0$ when $D<l_0/\beta$, otherwise $2{\beta}l/(2\beta-1)$. 

\begin{figure}
\begin{center}
\includegraphics[width=\linewidth, scale=0.5]{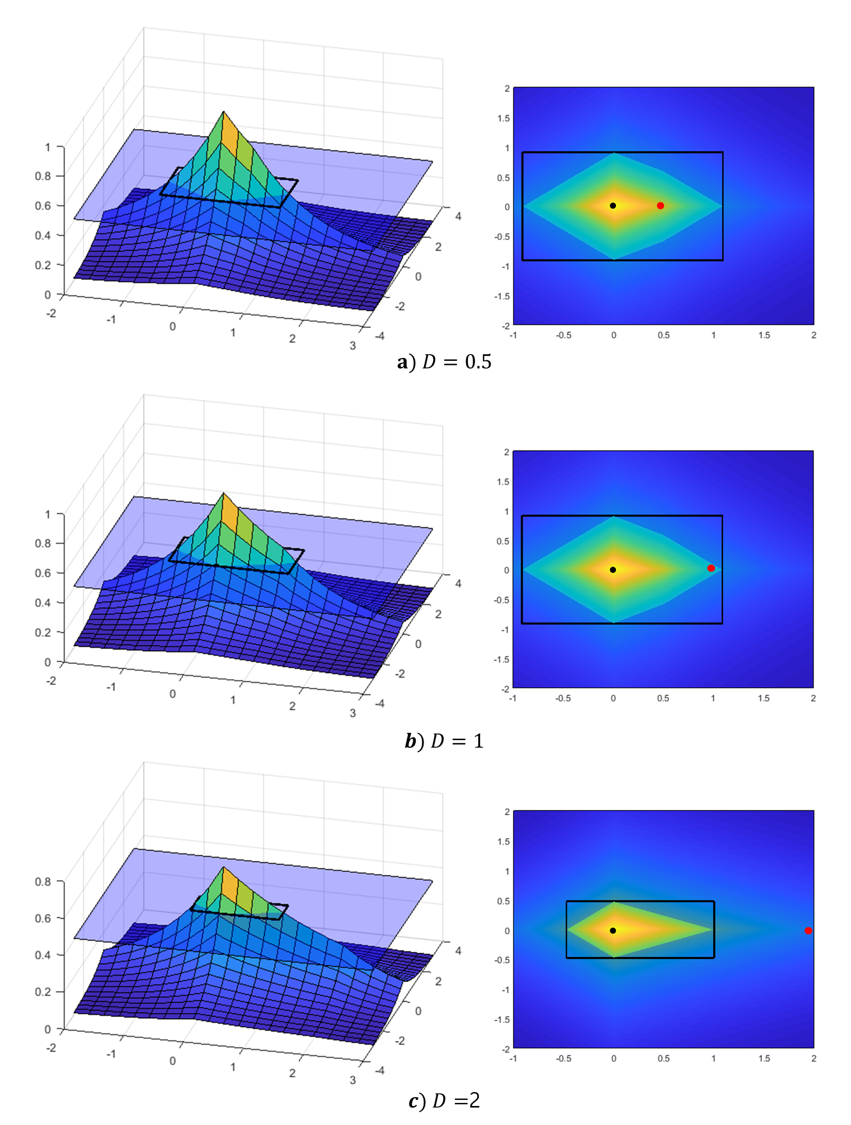}
\end{center}
   \caption{
   The examples of guidance distributions in Section~\ref{section:Guidance distribution model}: Each row shows a and top-view of guidance distribution, according to the distance $D$ between visual corresponding point (black dot) and guidance point (red dot).
   The black rectangular envelopes of the $c$-level set (intersection with transparent plane) determine the the grid as the black rectangle in top-view.
   }
\label{fig:Proposed_distribution}
\end{figure}

\subsection{Energy scaling}
\label{section:Energy scaling}

The most efficient way to obtain sufficient image statistics for correspondence is computing patch error as used in feature tracking algorithms. Unfortunately, descriptors in feature matching algorithm is not appropriate in sub-pixel precision. Thus, we utilize a root mean squared error function $e$ between a target image patch centered by $\boldsymbol{\mathrm{x}}$ and the reference image patch centered by the reference point $\boldsymbol{\mathrm{x}}^{(i)}$ as follows:
\begin{equation}
\label{eqn:1}
e(\boldsymbol{\mathrm{x}};\boldsymbol{\mathrm{x}}^{(i)})=\sqrt{\frac{1}{|G|}\sum_{\boldsymbol{\mathrm{d}}\in G} (\boldsymbol{\mathrm{I}}^{(i)}[\boldsymbol{\mathrm{x}}^{(i)}+\boldsymbol{\mathrm{d}}]-\boldsymbol{\mathrm{I}}^{(j)}[\boldsymbol{\mathrm{x}}+\boldsymbol{\mathrm{d}}])^{2}},
\end{equation}
where $\boldsymbol{\mathrm{I}}^{(k)}: \mathbb{R}^2 \to{} [0,1]$ is a function retrieving the image intensity at time $k$ and $G$ is a given set of displacement $\boldsymbol{\mathrm{d}}$. Note that there is no guarantee that the visual corresponding point is the minimum of the patch error.

Therefore, an error-based energy $\Psi_{\text{e}}$ for true corresponding point can be linearly approximated with the patch error as below:
\begin{equation}
\label{eqn:eng_error}
\Psi_{\text{e}}(\boldsymbol{\mathrm{x}};\boldsymbol{\mathrm{x}}^{(i)})=ke(\boldsymbol{\mathrm{x}};\boldsymbol{\mathrm{x}}^{(i)})+C,
\end{equation}
where $k$ is a scale factor and $C$ is a negligible constant. Thus, the conditional distribution inside a region $R$ where the approximation holds is determined by the energy $\Psi_{e}$, or set to 0 elsewhere, as below:
\begin{align}
\label{eqn:prob_error}
&p_k(\boldsymbol{\mathrm{x}}|\boldsymbol{\mathrm{x}}^{(i)})=\left.
\begin{cases}
\frac{\exp[-\Psi_{\text{e}}(\boldsymbol{\mathrm{x}})]}{\mathrm{Z}_k}, & \text{if } \boldsymbol{\mathrm{x}}\in R \\
0, & \text{else,} \\
\end{cases}\right. \\
&\mathrm{Z}_k=\int_{R}{\mathrm{exp}[-\Psi_{\text{e}}(\boldsymbol{\mathrm{x}})]\boldsymbol{\mathrm{dx}}},
\end{align}
where $\mathrm{Z}_k$ is a normalization factor over the region $R$.

In order to find the best fit $k^*$, rather than conventional suppression methods~\cite{nickels2002estimating}, we minimize Kullback–Leibler divergence $D_{\text{KL}}$ between the error distribution and the guidance distribution, in order to reflect the shape of guidance distribution. Using the possible correspondence sample points $\boldsymbol{\mathrm{X}}$, hopefully, inside the region, we can approximate it as below:
\begin{equation}
\begin{aligned}
\label{eqn:KL_1}
k^*= &\argmin_k D_{\text{KL}}(p_k(\boldsymbol{\mathrm{x}}|\boldsymbol{\mathrm{x}}^{(i)})||q(\boldsymbol{\mathrm{x}}|\boldsymbol{\mathrm{x}}_{v},\boldsymbol{\mathrm{x}}_{g})) \\
\approx & \argmin_k\sum_{\boldsymbol{\mathrm{x}}\in \boldsymbol{\mathrm{X}}} p_k(\boldsymbol{\mathrm{x}}|\boldsymbol{\mathrm{x}}^{(i)})\log\frac{p_k(\boldsymbol{\mathrm{x}}|\boldsymbol{\mathrm{x}}^{(i)})}{q(\boldsymbol{\mathrm{x}}|\boldsymbol{\mathrm{x}}_{v},\boldsymbol{\mathrm{x}}_{g})},
\end{aligned}
\end{equation}
However, the best fit $k^*$ cannot be determined due to the intractable normalization factor $\mathrm{Z}_k$. As a constraint, we regularize the probability at the visual correspondence to be equal to that of the guidance distribution as below:
\begin{equation}
\begin{aligned}
\label{eqn:norm_equal}
p_k(\boldsymbol{\mathrm{x}}_v|\boldsymbol{\mathrm{x}}^{(i)})=q(\boldsymbol{\mathrm{x}}_v|\boldsymbol{\mathrm{x}}_{v},\boldsymbol{\mathrm{x}}_{g}).
\end{aligned}
\end{equation}

By combining the Eq.~\ref{eqn:KL_1} and \ref{eqn:norm_equal}, the best fit $k^*$ is determined by a nonlinear equation as below:
\begin{equation}
\label{eqn:KL_2}
\frac{\partial}{\partial k} \sum_{\boldsymbol{\mathrm{x}}\in \boldsymbol{\mathrm{X}}} p_k(\boldsymbol{\mathrm{x}}|\boldsymbol{\mathrm{x}}^{(i)})\log\frac{p_k(\boldsymbol{\mathrm{x}}|\boldsymbol{\mathrm{x}}^{(i)})}{q(\boldsymbol{\mathrm{x}}|\boldsymbol{\mathrm{x}}_{v},\boldsymbol{\mathrm{x}}_{g})}=0.
\end{equation}
We can find the solution by a bisection method, because the left-hand side of the equation is continuous function with an unique root~\cite{brent2013algorithms}.

\subsection{Point-level uncertainty marginalization}
\label{section:Point-level uncertainty marginalization}

\begin{figure*}[t]\centering
  \includegraphics[height=9cm]{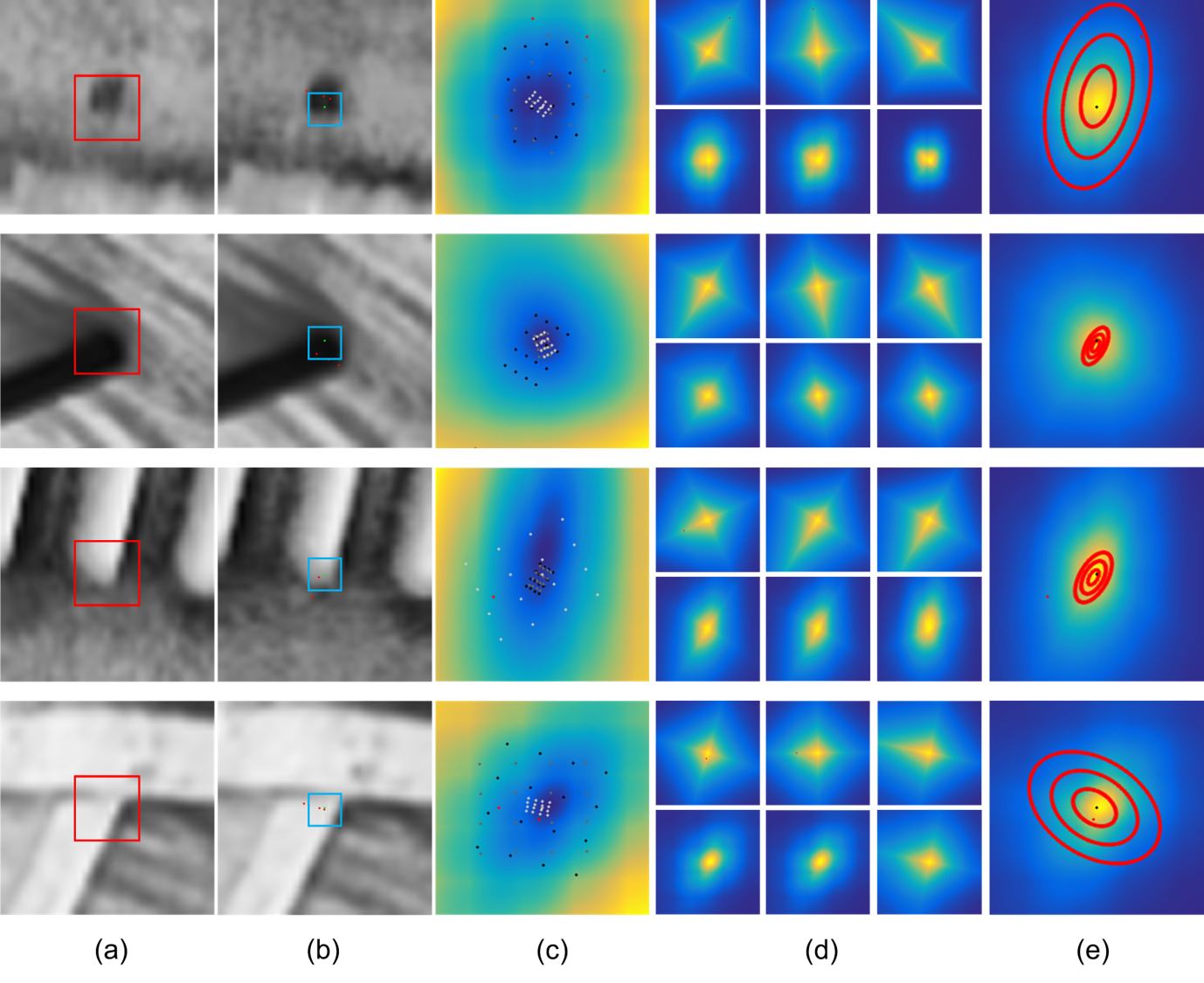}
  \caption{
  Examples of uncertainty estimation procedure in various image patches: (a) Image patch around the reference point. The red box represents the patch used to calculate the patch error. (b) Image patch around the visual corresponding point. The blue box represents the region of (c), (d), (e). The green point is the visual corresponding point and red points are the guidance points. (c) Patch errors around the visual corresponding point. The white, gray, and black points are the samples of the grid in Section~\ref{section:Inertial guidance sampling}. (d) The top row is the guidance distribution in Section~\ref{section:Guidance distribution model}. The bottom row represents the combined distribution of Eq.~\ref{eqn:combined_energy}. (e)
  The marginal distribution in Eq.~\ref{eqn:marginalized_distribution}, and the red ellipses represent the Gaussian distribution approximation in Section~\ref{section:Point-level uncertainty marginalization}.
  For the visualization of the above distributions except the guidance distribution, we fill the dense patch errors unobservable at the actual procedures. 
  }
  \label{fig:whole_procerdure}
\end{figure*}

Each of the scaled error-based energy functions from multiple guidance points represents its respective hypothesis of a true corresponding point. Thus, the point-level uncertainty of the true corresponding point can be marginalized with the guidance points to a multi-modal distribution as below:
\begin{align}
\label{eqn:point_level_uncertainty}
&p(\boldsymbol{\mathrm{x}}^{(j)} | \boldsymbol{\mathrm{x}}^{(i)}, \boldsymbol{\mathrm{x}}_{v}, \boldsymbol{\mathrm{R}}_{I}^{(i,j)}, \boldsymbol{\mathrm{p}}_I^{(i,j)})\\
\label{eqn:marginalized_distribution}
&=\int{p(\boldsymbol{\mathrm{x}}^{(j)}|\boldsymbol{\mathrm{x}}^{(i)},\boldsymbol{\mathrm{x}}_{v},\boldsymbol{\mathrm{x}}_{g})p(\boldsymbol{\mathrm{x}}_{g}|\boldsymbol{\mathrm{R}}_{I}^{(i,j)}, \boldsymbol{\mathrm{p}}_{I}^{(i,j)})d\boldsymbol{\mathrm{x}}_{g}}
\end{align}
where $p(\boldsymbol{\mathrm{x}}^{(j)}|\boldsymbol{\mathrm{x}}^{(i)},\boldsymbol{\mathrm{x}}_{v},\boldsymbol{\mathrm{x}}_{g})$ is a conditional distribution determined by the weighted mean of the scaled error-based energy function and the guided function.
\begin{equation}
p(\boldsymbol{\mathrm{x}}|\boldsymbol{\mathrm{x}}^{(i)},\boldsymbol{\mathrm{x}}_{v},\boldsymbol{\mathrm{x}}_{g})=\frac{1}{Z_{k^{*}}^{(1-\lambda)}Z_{\alpha,\beta}^{\lambda}}\exp[(1-\lambda)\Psi_{\text{e}}+\lambda\psi_{\text{d}}],
\label{eqn:combined_energy}
\end{equation}
where $\lambda$ is a weight parameter automatically adjusted by the minimum energy values of the energy functions.

Though the marginalized distribution is nearly uni-modal at the visual corresponding point, the mean and covariance matrix is computed in multi-modal sense as below:
\begin{align}
&\boldsymbol{\mu}=\frac{1}{\sum_n{|\boldsymbol{\mathrm{X}}_n}|}\sum_{n}\sum_{\boldsymbol{\mathrm{x}} \in \boldsymbol{\mathrm{X}}_n}{\boldsymbol{\mathrm{x}}p(\boldsymbol{\mathrm{x}} |\boldsymbol{\mathrm{x}}_{g}^{n})}, \label{eqn:sample mean}\\
&\boldsymbol{\Sigma}=\frac{1}{\sum_n{|\boldsymbol{\mathrm{X}}_n}|}\sum_{n}\sum_{\boldsymbol{\mathrm{x}} \in \boldsymbol{\mathrm{X}}_n}{(\boldsymbol{\mathrm{x}}-\boldsymbol{\mu})(\boldsymbol{\mathrm{x}}-\boldsymbol{\mu})^Tp(\boldsymbol{\mathrm{x}} |\boldsymbol{\mathrm{x}}_{g}^{n})}, \label{eqn:sample covariance}
\end{align}
where $\boldsymbol{\mathrm{x}}_{g}^{n}$ and $\boldsymbol{\mathrm{X}}_n$ are the $n$-th sample of the guidance point and the $n$-th sample set of possible corresponding points, respectively. The example of each step of uncertainty estimation is shown in Fig.~\ref{fig:whole_procerdure}.

Also, there would be a number of correspondences in multiple image pairs in a sequence of consecutive images. The correspondences are established not only between two consecutive images but also from the very first one~\cite{qin2018vins} or the most representative one~\cite{rublee2011orb}. However, comparing two image features from significantly different viewpoints is not reliable. Therefore, we predict the uncertainty between consecutive images and propagate the uncertainty from the first detection to the latest observation. This propagation method samples multiple reference points from the propagated uncertainty, while the very first uncertainty would be chosen without the propagation. Thus, the point-level uncertainty should be marginalized by the uncertain reference point. Additionally, we normalize the covariance matrices in the same image so that the averaged determinants in each image should be equal, in order to prevent the overall uncertainty from nearly removing uncertain images in a window and balance with other measurement uncertainties.

\section{Experiments}

We validate our uncertainty estimation by using VINS-mono~\cite{qin2018vins} which is one of the most efficient VIO/SLAM algorithms. We apply our uncertainty estimation method to both the feature tracking algorithm of the front-end and the bundle adjustment of the back-end. Thus, the locations of tracked points are modified by the mean of the uncertainty and the visual measurement residual defined by Mahalanobis distance is re-calculated with the covariance matrix of the uncertainty. Also, the uncertainty is applied to the triangulation for depth initialization.

We also investigate two baseline uncertainty estimation methods: The first one is the method used in VINS-mono, where the covariance matrix is a scaled identity matrix. The scale of the covariance matrix would be set to control the relative importance between visual residual and inertial residual. Thus, we constrain our covariance matrix to have the similar scale in average. The second baseline method is a modified version of ~\cite{nickels2002estimating} where the sum of squared errors is replaced with more robust root mean squared error, and resulting covariance matrix is also normalized like the first method.

For evaluation, we compute relative pose errors~\cite{geiger2012we} for 10 seconds, since we focus mainly on local odometry. We evaluate our method and the two baseline methods at TUM VI~\cite{schubert2018tum} dataset and ZJU~\cite{jinyu2019survey} dataset.

\subsection{CVG-ZJU dataset}
CVG-ZJU dataset~\cite{jinyu2019survey} is an indoor smart-phone AR dataset in which a person watching a phone walks around a room. It contains A and B sequences of 8 sub-sequences. We evaluate on 7 A sequences, because the rest of the sequences have extreme situations that makes the VINS-mono fail often.

\begin{figure}[t]
\begin{center}
\includegraphics[width=\linewidth, scale=0.7]{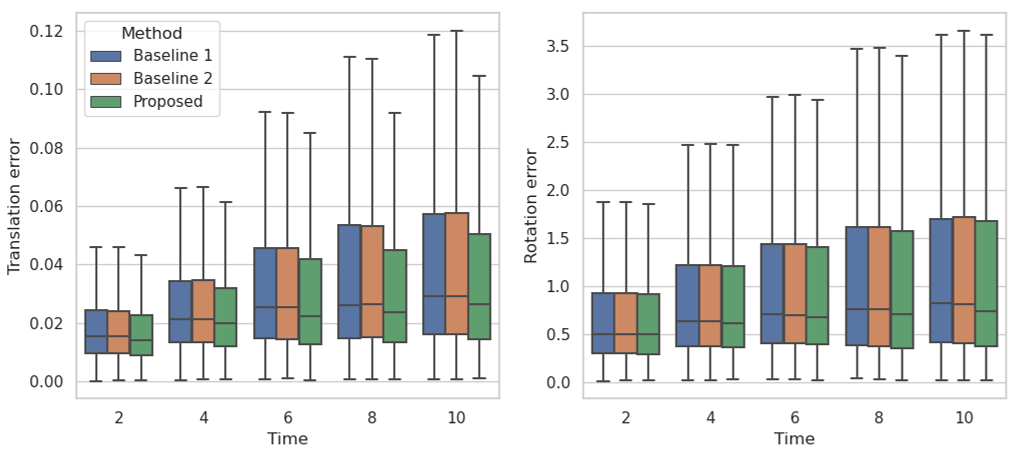}
\end{center}
   \caption{The error statistics of translation (mm) and rotation (degree) of 'A' sequences ZJU dataset (except A5), for time difference from 2 to 10 seconds.}
\label{fig:ZJU_2to10}
\end{figure}

In Table~\ref{table:zju_result}, the proposed method shows the promising result, showing that the average translation and rotation error are reduced by 7.6\% and 3.3\% respectively. The result also demonstrates that relatively short-term errors over 10 seconds from the feature tracking are reduced by our uncertainty estimation. In Fig.~\ref{fig:ZJU_2to10}, the statistics of translation and rotation errors from 2 to 10 seconds are shown in boxplot. The result shows that our uncertainty estimation also decrease variance and maximum value of the error as well as mean of the error.

\subsection{TUM VI dataset}
TUM VI~\cite{schubert2018tum} is a visual-inertial dataset which consists of 28 sequences collected with a person walking with a fish-eye camera in hands. Among the sequences, we use the 6 sequences taken in a room, because they have the ground truth pose. 
The result of TUM VI dataset is shown in Table~\ref{table:tum_result}. While the rotation error is similar to that of two baseline methods, the average translation error is reduced about 1.7\% while the rotation error is similar with baseline methods. The result shows that the accumulated translation error in short term can be reduced by the uncertainty estimation.

\begin{table}
\centering
\begin{adjustbox}{width=\columnwidth,center}
\small{
\begin{tabular}{*7c}
\toprule[1.5pt]
Sequence & Baseline 1 & Baseline 2 & Proposed\\
\midrule
room 1 & 41.72 / 1.040 & 39.99 / 1.052 & 37.21 / 0.945 \\
room 2 & 79.57 / 1.184 & 78.78 / 1.188 & 80.92 / 1.300 \\
room 3 & 71.72 / 1.184 & 71.29 / 1.178 & 66.21 / 1.141 \\
room 4 & 30.99 / 0.545 & 30.14 / 0.544 & 36.37 / 0.549 \\
room 5 & 48.39 / 0.519 & 48.63 / 0.506 & 44.16 / 0.543 \\
room 6 & 21.51 / 0.466 & 24.05 / 0.455 & 30.91 / 0.470 \\
Avg.   & 54.64 / 0.898 & 54.04 / 0.896 & 53.15 / 0.899 \\
\bottomrule[1.5pt]
\end{tabular}
}
\end{adjustbox}
\caption{Translation (mm) errors and rotation (degree) errors of room sequences in TUM VI benchmark.}
\label{table:tum_result}
\end{table}

\subsection{Ablation study}
In the sampling process of guidance point in Section~\ref{section:Inertial guidance sampling}, the depth is sampled from a Gaussian distribution whose mean is the estimates of the VIO. For the points of which depth are not available, we use triangulation using two points to estimate the depth. 

To validate the sampling process of guidance point, in Table~\ref{table:ablation_g_sampling}, we compare the various methods with changing the number of points and changing the sampling method of candidate points. One of the candidate points is the nearest point on the epipolar line from the visual corresponding point (w/ epi). The point can be the one of the candidate of the guidance point, when the depth is inaccurate. The 3 variations contain at least one guidance point from the mean depth. The result shows that the one sample is vulnerable to error of the depth estimates. The point from epipolar line can be a candidate point, but has little effect on the result.

In Section~\ref{section:Possible correspondence sampling}, we sample the patch error on the grid. In Table~\ref{table:ablation_g_sampling}, we investigate effect of the parameter $l_0$ in Section~\ref{section:Possible correspondence sampling}. $2l_0$ represents the default width of square grid when visual corresponding point and guidance point are same. To validate the effect of $l_0$, we fix the distance between the points in the grid to maintain the sparsity of the grid. The number of points ($N_e$) is also changed to maintain the sparsity. The result shows that there are some trade-off about the grid size, because the point far from the visual corresponding point has little relationship with the true corresponding point.
The further experiments and implementation details are represented in supplementary materials.

\begin{table}
\centering
\begin{adjustbox}{width=\columnwidth,center}
\small{
\begin{tabular}{*7c}
\toprule
Sequence & Baseline 1 & Baseline 2 & Proposed\\
\midrule
A0 & 140.42 / 3.788 & 139.98 / 3.784 & 142.15 / 3.764 \\
A1 & 52.44 / 1.191 & 52.3 / 1.184 & 42.61 / 1.029 \\
A2 & 56.06 / 1.910 & 56.12 / 1.910 & 53.18 / 1.871 \\
A3 & 25.58 / 1.305 & 25.00 / 1.301 & 22.50 / 1.346 \\
A4 & 19.63 / 0.409 & 20.15 / 0.406 & 21.47 / 0.374 \\
A6 & 23.09 / 0.978 & 23.72 / 0.957 & 20.16 / 1.001 \\
A7 & 20.65 / 0.395 & 20.34 / 0.388 & 15.76 / 0.372 \\
Avg. & 45.56 / 1.275 & 45.60 / 1.268 & 42.11 / 1.225 \\
\bottomrule
\end{tabular}
}
\end{adjustbox}
\caption{Translation errors (mm) and rotation (degree) errors of A sequences in ZJU dataset, excluding A5 sequence.}
\label{table:zju_result}
\end{table}
 
\begin{table}
\centering
 \label{tab:title} 
\begin{adjustbox}{width=\columnwidth,center}
\small{
\begin{tabular}{*6c}
\multicolumn{4}{c}{Guidance point sampling} \\
\toprule
Number of guidance &  \#1 & \#3 & \#3 (w/ epi)\\
\midrule
 mm / deg. &  42.71 / 1.314 & 42.11 / 1.247 & 42.53 / 1.247  \\
\bottomrule
\end{tabular}
}
\end{adjustbox}

\begin{adjustbox}{width=\columnwidth,center}
\small{
\begin{tabular}{*6c}
\multicolumn{4}{c}{} \\
\multicolumn{4}{c}{Patch error sampling} \\
\toprule
$N$ / $l_0$ &  9 / 0.66 &  16 / 1 & 25 / 1.33\\
\midrule
mm / deg. & 43.07 / 1.241 & 42.11 / 1.225 & 42.24 / 1.225 \\
\bottomrule
\end{tabular}
}
\end{adjustbox}

\caption{The results of ablation studies averaged over A sequences of ZJU dataset (except A5): The top table shows the comparison with variations of guidance point sampling in Section~\ref{section:Inertial guidance sampling}. The second table shows the comparison over the number of sample point and grid size. The parameter $N$ is the number of points in grid, and $l_0$ is the parameter about default grid size in Eq.~\ref{eqn:gridset}.}
\label{table:ablation_g_sampling}
\end{table}

\section{Conclusion}
In this paper, we have proposed a novel uncertainty estimation method for feature correspondence in visual-inertial odometry/SLAM. The proposed method predicts the true corresponding point using robust but probabilistic guidance points re-projected from the inertial pose and uncertain depth. Each guidance point yields distance-based energy about the prior distribution of visual corresponding point drift. Then, multiple image-based energy functions are properly scaled and combined to predict the distribution of the true corresponding point. The predicted distribution can be applied to both the front-end and back-end of existing feature-based odometry/SLAM algorithms. The experimental results show that our method reduces relative pose error in most sequences of which feature correspondences are tolerably uncertain. Hopefully, such contribution may boost overall competitiveness of feature-based approaches by granting robustness similar to the benefit of direct/dense approaches.  We plan to improve our method by utilizing more informative guidance than multiple re-projections and more specialized image statistics than patch errors.

\captionsetup[table]{skip=10pt}
\captionsetup[figure]{skip=0pt}
\renewcommand\thesection{\Alph{section}}
\renewcommand\thesubsection{\thesection.\arabic{subsection}}

\newpage

\newcommand{\beginsupplement}{%
        \setcounter{table}{0}
        \renewcommand{\thetable}{S\arabic{table}}%
        \setcounter{figure}{0}
        \renewcommand{\thefigure}{S\arabic{figure}}%
     }
\beginsupplement
\begin{center}
\textbf{\Large Supplementary Materials}
\end{center}
\setcounter{equation}{0}
\setcounter{figure}{0}
\setcounter{table}{0}
\setcounter{section}{0}
\makeatletter
\renewcommand{\theequation}{S\arabic{equation}}
\renewcommand{\thefigure}{S\arabic{figure}}


\section{Experimental details}
\label{Experimental details}
To sample a guidance point in Section \ref{section:Inertial guidance sampling}, we set the standard deviation of depth distribution to 10\% of the estimated depth. Also, the guidance point is clipped by the maximum distance $D_\text{max}$ from the visual corresponding point, to prevent the size of the grid, $l=l_0-(1-\beta)D$ in Section \ref{section:Possible correspondence sampling}, from being negative.

We choose the parameters of guidance distribution $q(\boldsymbol{\mathrm{x}}|\boldsymbol{\mathrm{x}}_{v},\boldsymbol{\mathrm{x}}_{g})$ in Eq.~\ref{eqn:8}. Regardless of the sampled guidance point, $\alpha$ is set as 0.89 in CVG-ZJU~\cite{jinyu2019survey} and 0.53 in TUM VI dataset~\cite{schubert2018tum}. Then, $\beta$ is calculated by Eq.~\ref{eqn:param_r} of which $r$ linearly increases from 1 to 3 in proportion to the distance between the guidance point and the visual corresponding point. 

In Section \ref{section:Possible correspondence sampling}, we set the default size of the grid $l_0$ to one pixel in CVG-ZJU and half pixels in TUM VI dataset, respectively. This is because the IMU in TUM VI dataset has lower accuracy and we reduce the importance of the guidance point. The visual corresponding point is also sampled in addition to 16 points on the grid. The patch error is calculated with a 10 by 10 image patch.

We perform experiments on AMD Ryzen 9 3900x 12-core processor running at 4.6 GHz with 32GB of memory. We use the public codes of VINS-mono~\cite{qin2018vins} in \cite{vinsmonocode} on Ubuntu 18.04 and ROS Melodic.

\section{Variances of experimental results}
\label{Variance of experimental results}
All experiments in the main manuscript and the supplementary materials are performed four times. Table \ref{table:std_check} shows the standard deviations for all sequences of CVG-ZJU and TUM VI datasets. The standard deviations are computed on the four trial results of each experiment.
The average result of each dataset shows that the improvement is larger than 2 to 8 times of the standard deviation except the rotation error in TUM VI dataset. The variance is caused by multiple sources of randomness such as RANSAC~\cite{fischler1981random}, the delay of ROS system and so on. Our uncertainty estimation method also inherits randomness from the sampling techniques, but it does not increase overall variance as shown in Table \ref{table:std_check}. 

\begin{table}
\counterwithin{table}{section}
\centering

\begin{adjustbox}{width=\linewidth,center}
\small{
\begin{tabular}{*7c}
\multicolumn{3}{c}{CVG-ZJU dataset} \\
\toprule
Seq. & Baseline 1 & Proposed\\
\midrule
A0 & 140.42 $\pm$0.22 / 3.788 $\pm$0.008 & 142.15 $\pm$0.84 / 3.764 $\pm$0.004\\
A1 & 52.44 $\pm$1.28 / 1.191 $\pm$0.029 & 42.61 $\pm$1.27 / 1.029 $\pm$0.014\\
A2 & 56.06 $\pm$0.00 / 1.910 $\pm$0.000 & 53.18 $\pm$1.35 / 1.871 $\pm$0.010 \\
A3 & 25.58 $\pm$0.54 / 1.305 $\pm$0.002 & 22.50 $\pm$1.20 / 1.346 $\pm$0.018 \\
A4 & 19.63 $\pm$1.26 / 0.409 $\pm$0.019 & 21.47 $\pm$0.92 / 0.374 $\pm$0.008 \\
A6 & 23.09 $\pm$0.15 / 0.978 $\pm$0.017 & 20.16 $\pm$0.61 / 1.001 $\pm$0.012 \\
A7 & 20.65 $\pm$0.01 / 0.395 $\pm$0.006 & 15.76 $\pm$0.79 / 0.372 $\pm$0.012 \\
Avg. & 45.56 $\pm$0.40 / 1.275 $\pm$0.006 & 42.11 $\pm$0.68 / 1.225 $\pm$0.006 \\
\bottomrule
\end{tabular}
}
\end{adjustbox}

\begin{adjustbox}{width=\linewidth,center}
\small{
\begin{tabular}{*7c}
\multicolumn{3}{c}{} \\
\multicolumn{3}{c}{TUM VI dataset} \\
\toprule
Seq. & Baseline 1 & Proposed\\
\midrule
room1 & 41.72 $\pm$1.60 / 1.040 $\pm$0.013 & 37.21 $\pm$0.67 / 0.945 $\pm$0.007\\
room2 & 79.57 $\pm$0.94 / 1.184 $\pm$0.006 & 80.92 $\pm$1.41 / 1.300 $\pm$0.006 \\
room3 & 71.72 $\pm$0.34 / 1.184 $\pm$0.007 & 66.21 $\pm$0.82 / 1.141 $\pm$0.008 \\
room4 & 30.99 $\pm$1.26 / 0.545 $\pm$0.002 & 36.37 $\pm$0.41 / 0.549 $\pm$0.033 \\
room5 & 48.39 $\pm$2.95 / 0.519 $\pm$0.013 & 44.16 $\pm$1.24 / 0.543 $\pm$0.017 \\
room6 & 21.51 $\pm$4.82 / 0.466 $\pm$0.033 & 30.91 $\pm$2.50 / 0.470 $\pm$0.017 \\
Avg. & 54.64 $\pm$0.60 / 0.898 $\pm$0.003 & 53.15 $\pm$0.27 / 0.899 $\pm$0.007 \\
\bottomrule
\end{tabular}
}
\end{adjustbox}
\caption{Translation (mm) and rotation (degree) errors and their corresponding standard deviations of CVG-ZJU and TUM VI datasets.}
\label{table:std_check}
\end{table}

\section{Various incorporation methods}
\label{Various incorporation methods}

\begin{table}
\counterwithin{table}{section}
\centering
\begin{adjustbox}{width=7cm,center}
\small{
\begin{tabular}{cc|cc}
\toprule
Seq. & Visual mean & Single sample & w/ visual\\
\midrule
A0 & 141.04 / 3.784 & 142.15 / 3.764 & 142.22 / 3.766\\
A1 & 50.23 / 1.176 &  42.61 / 1.029 & 43.49 / 1.019\\
A2 & 56.69 / 1.913 &  53.18 / 1.871 & 52.31 / 1.859\\
A3 & 24.23 / 1.306 &  22.50 / 1.346 & 22.06 / 1.356\\
A4 & 20.36 / 0.411 &  21.47 / 0.374 & 21.15 / 0.375\\
A6 & 23.34 / 0.970 &  20.16 / 1.001 & 19.85 / 0.979\\
A7 & 20.39 / 0.394 &  15.76 / 0.372 & 17.07 / 0.381\\
Avg. & 45.25 / 1.270 & 42.11 / 1.225 & 42.26 / 1.220\\
\bottomrule
\end{tabular}
}
\end{adjustbox}
\caption{Translation (mm) and rotation (degree) errors of variations of the proposed method. The first column shows the result of maintaining the visual corresponding point without using the mean of the point-level uncertainty. The second and third columns present the results of sampling the reference point in two ways. The same sequences of CVG-ZJU dataset are used as in Table \ref{table:std_check}.}
\label{table:drift_result}
\end{table}

\begin{table}
\counterwithin{table}{section}
\centering
\begin{adjustbox}{width=5cm,center}
\small{
\begin{tabular}{*7c}
\toprule
Weight & Baseline 1 & Proposed\\
\midrule
0.05 & 48.19 / 1.300 & 49.03 / 1.275\\
0.1 & 48.07 / 1.301 & 48.06 / 1.242 \\
1.0 & 44.97 / 1.269 & 42.60 / 1.222 \\
5.0 & 45.74 / 1.274 & 42.50 / 1.241\\
10.0 & 45.56 / 1.275 & 42.11 / 1.225\\
\bottomrule
\end{tabular}
}
\end{adjustbox}
\caption{Average translation (mm) and rotation (degree) errors of each method with changing the weight of the residual of visual measurements. The same sequences of CVG-ZJU dataset are used as in Table \ref{table:drift_result}}
\label{table:g_scale_result}
\end{table}

We also perform experiments of several variations of incorporating our method into VINS-mono. Firstly, rather than correcting the actual corresponding point to the mean of the point-level uncertainty described in our approach, we maintain the visual corresponding point as in the baseline method. As shown in ``Visual mean" of Table \ref{table:drift_result}, if we use the visual corresponding point without correction, the performance is decreased to that of the baseline method.

Secondly, in Section \ref{section:Point-level uncertainty marginalization}, we propagate the uncertainty by multiple samples of reference point using the former point-level uncertainty. However, in order to avoid burdensome feature tracking of every samples, we alternatively average patch errors from multiple reference patches. In Table \ref{table:drift_result}, ``Single sample" represents using the mean of the former point-level uncertainty, and ``w/ visual" represents using both the mean and the former visual corresponding point in reference image. The results show that simple propagation using single sample is sufficient to propagate uncertainty. 

In addition, there is a parameter to adjust the weight of visual measurement residual in the cost function of VIO ~\cite{qin2018vins}. As in Table \ref{table:g_scale_result}, our approach yields better performance for larger weights on the visual measurement, while the baseline method performs worse for larger values of weight. We can see that our uncertainty estimation approach makes the visual measurement residual more reliable than the baseline method.

\section{Results on EuRoC dataset}
\label{Result on EuRoC dataset}

\begin{table}
\counterwithin{table}{section}
\centering
\begin{adjustbox}{width=6cm,center}
\small{
\begin{tabular}{*7c}
\toprule
Seq. & Baseline 1 & Proposed\\
\midrule
     MH 01 easy & 334.11 / 5.189 & 337.33 / 5.179\\
     MH 02 easy & 210.91 / 2.796 & 208.39 / 2.799 \\
   MH 03 medium & 183.08 / 0.520 & 177.29 / 0.449\\
MH 04 difficult & 263.48 / 0.507 & 252.47 / 0.504 \\
MH 05 difficult & 197.74 / 0.342 & 204.97 / 0.349\\
     V1 01 easy & 253.71 / 2.861 & 252.54 / 2.868\\
   V1 02 medium & 129.22 / 0.863 & 127.67 / 0.858 \\
V1 03 difficult & 130.80 / 1.205 & 131.70 / 1.161\\
     V2 01 easy &  70.61 / 0.736 & 70.41/ 0.832\\
   V2 02 medium &  95.35 / 1.187 & 97.53 / 1.173 \\
V2 03 difficult & 168.83 / 1.400 & 170.17 / 1.344\\
           Avg. & 198.01 / 1.934 & 197.47 / 1.925\\
\bottomrule
\end{tabular}
}
\end{adjustbox}
\caption{Translation (mm) and rotation (degree) errors of EuRoC dataset.}
\label{table:result of EuRoC dataset}
\end{table}

We test the proposed approach in EuRoC dataset which was taken from a micro aerial vehicle (MAV) containing larger translational motion than human motion and lots of noise in IMU measurements due to the vibration of motors~\cite{wei2018rapid}. These two factors can cause less reliable guidance points, resulting in smaller improvement of performance compared with other datasets, as shown in Table~\ref{table:result of EuRoC dataset}.

\end{document}